\documentclass[lettersize,journal]{IEEEtran}
\usepackage{amsmath,amsfonts}
\usepackage{algorithmic}
\usepackage{algorithm}
\usepackage{array}
\usepackage[caption=false,font=footnotesize]{subfig}
\usepackage{textcomp}
\usepackage{stfloats}
\usepackage{url}
\usepackage{verbatim}
\usepackage{graphicx}
\usepackage{xcolor}
\usepackage{cite}
\begin{document}

\title{Exp-Graph: How Connections Learn Facial
Attributes in Graph-based Expression Recognition}

\author{
\IEEEauthorblockN{Nandani Sharma, Dinesh Singh}\\
\IEEEauthorblockA{\textit{Vision Intelligence and Machine Learning (VIML) Group} \\
\textit{School of Computing and Electrical Engineering, Indian Institute of Technology Mandi, India.}\\
d22180@students.iitmandi.ac.in, dineshsingh@iitmandi.ac.in}

\thanks{This paper was produced by the IEEE Publication Technology Group.}
\thanks{Manuscript received April XX, 2025; revised August XX, 2025.}}

\markboth{Journal of \LaTeX\ Class Files,~Vol.~14, No.~x, August~2025}%
{Shell \MakeLowercase{\textit{et al.}}: A Sample Article Using IEEEtran.cls for IEEE Journals}

\maketitle

\begin{abstract}
    Facial expression recognition is crucial for human-computer interaction applications such as face animation, video surveillance, affective computing, medical analysis, etc. Since the structure of facial attributes varies with facial expressions, incorporating structural information into facial attributes is essential for facial expression recognition. In this paper, we propose Exp-Graph, a novel framework designed to represent the structural relationships among facial attributes using graph-based modeling for facial expression recognition. For facial attributes graph representation, facial landmarks are used as the graph's vertices. At the same time, the edges are determined based on the proximity of the facial landmark and the similarity of the local appearance of the facial attributes encoded using the vision transformer. Additionally, graph convolutional networks are utilized to capture and integrate these structural dependencies into the encoding of facial attributes, thereby enhancing the accuracy of expression recognition. Thus, Exp-Graph learns from the facial attribute graphs highly expressive semantic representations. On the other hand, the vision transformer and graph convolutional blocks help the framework exploit the local and global dependencies among the facial attributes that are essential for the recognition of facial expressions. We conducted comprehensive evaluations of the proposed Exp-Graph model on three benchmark datasets: Oulu-CASIA, eNTERFACE05, and AFEW. The model achieved recognition accuracies of 98.09\%, 79.01\%, and 56.39\%, respectively. These results indicate that Exp-Graph maintains strong generalization capabilities across both controlled laboratory settings and real-world, unconstrained environments, underscoring its effectiveness for practical facial expression recognition applications.
\end{abstract}

\begin{IEEEkeywords}
Facial expression recognition, graph convolutional networks, and vision transformer.
\end{IEEEkeywords}

\section{Introduction}\label{sec:intro}
\begin{figure}[!t]
        \centering
        \includegraphics[width=1\linewidth, keepaspectratio]{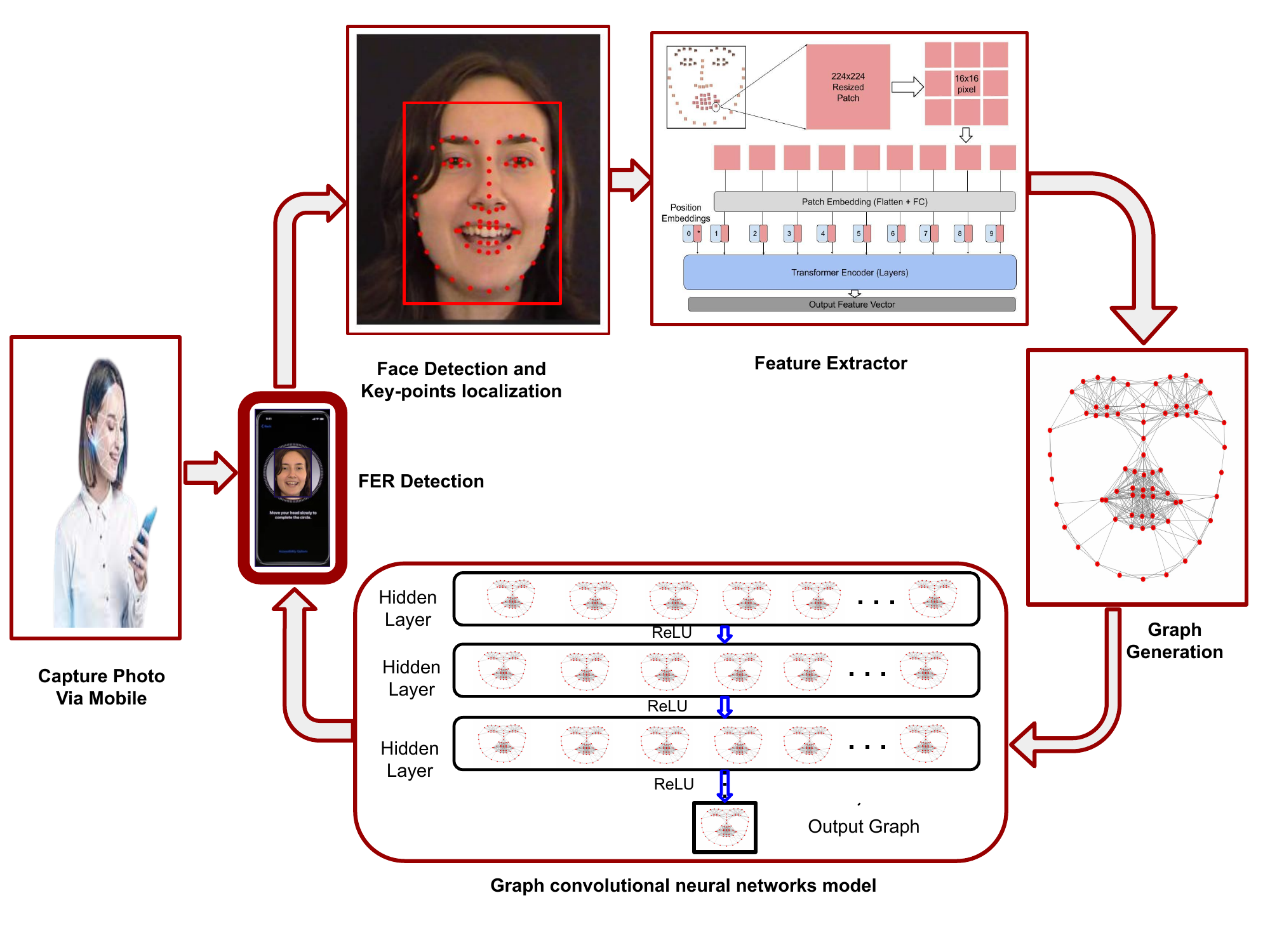}
        \caption{System architecture of facial expression recognition using Exp-Graph framework. [Best shown in color]}
        \label{fig:system_architecture}
    \end{figure}

\IEEEPARstart{F}{acial} expression recognition (FER) has garnered significant attention in computer vision research over the last few decades because of its critical role in enabling computers to comprehend human emotions and engage in human-to-human communication as shown in Fig.~\ref{fig:system_architecture}. However, their success lies in learning robust and discriminative representations of the expressions ( such as AN: anger, DI: disgust, FE: fear, HA: happy, SA: sad, SU: surprise, NE: neutral) from the facial images that are invariant to variations in the angle of viewpoint, lighting conditions, and head postures. Thus, devising a suitable feature representation is the primary objective of facial expression recognition. In the early research, successful hand-crafted features, such as Gabor wavelets~\cite{xie2006gabor}, local binary pattern (LBP)~\cite{zhao2007dynamic,shan2009facial}, histogram of oriented gradients (HoG)~\cite{dhall2011emotion,liew2015facial,chen2016facial,wang2020facial}, etc., were used to represent the facial expressions. However, these methods fall short of adequate recognition in more complicated real-world situations because of their low semantic correlation with facial expressions.

    Since deep learning techniques have rapidly advanced in the past decade, numerous attempts have been made to investigate discriminative representations for various recognition tasks. Deep models for visual representation, especially convolutional neural networks (CNNs) and vision transformers (ViTs), have shown promising results in various real-world applications because they can effectively learn discriminatory feature representations from visual observations~\cite{lee2019context,zhao2022spatial, mittal2020emoticon, minaee2021deep}. Also, several studies have used CNNs to improve semantic representations of facial expressions and demonstrated good results in identifying human emotions~\cite{wen2016discriminative,li2017reliable,cai2018island,li2020facial,zhao2021deep}. However, they mostly depend on the appearance only and cannot exploit the deep structure for the problems where the training data is limited~\cite{singh2017graph,huang2022graphlime,singh2023graph}. 
    
    Vision transformers~\cite{dosovitskiy2020image} capture global context by implementing self-attention processes; they are becoming a more popular choice for visual feature extraction compared to standard deep neural networks (DNNs) like CNNs. Vision transformers can analyze an image as a sequence of patches, allowing for greater input size flexibility and better scalability with large-scale datasets than CNNs, which concentrate on local patterns only. 
    Additionally, ViTs are more generalizable because of their excellent transfer learning capabilities and less inductive bias. Vision transformers are increasingly favoured for visual feature extraction over traditional CNNs due to their ability to capture global context via self-attention, offering better scalability and input flexibility. Unlike CNNs, which focus on local patterns and are sensitive to variations like lighting and occlusions, ViTs can generalize better and enhance transfer learning.

    \begin{figure}[!t]
        \centering
        \includegraphics[width=1\linewidth, keepaspectratio]{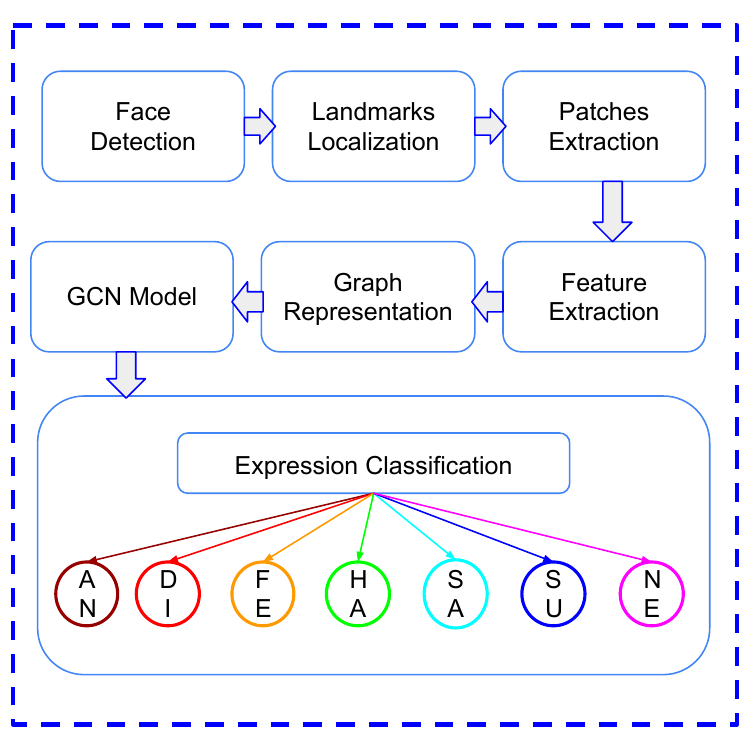}
        \caption{General pipeline of our system.}
        \label{fig:pipeline}
    \end{figure}
  
    This research explores using ViTs for facial expression recognition to address the limitations of CNNs, aiming to improve accuracy and reliability in recognizing complex facial expressions~\cite{li2023facial,liu2023facial}. Attention methods like graph attention (GAT)~\cite{velivckovic2017graph} and ViT~\cite{wu2020visual} at both geometry and appearance levels are used to improve FER performance.
    Vision transformers excel in capturing global features for image recognition, including FER, but they struggle with local feature extraction and require large datasets~\cite{dosovitskiy2020image,ma2021facial}. Graph convolutional networks (GCNs) address these limitations by enhancing local feature detection and improving data efficiency, making ViTs more effective for FER by capturing subtle facial details~\cite{zhang2020geometry, zhao2022spatial, luo2024hypergraph}. Integrating GCNs with ViTs offers a balanced solution, combining global and local feature learning for improved performance. Since facial expression highly depends on the relative change in the structure of facial attributes, incorporating graph structures of the facial attributes can significantly improve the facial expression representation, mainly when relying on transfer learning for visual encoding. However, traditional DNNs also struggle with non-Euclidean data, such as graphs, where relationships between data points are complex and irregular. Also, extensive feature engineering is required to capture complex relationships between data points. Scalability is another challenge for DNNs when dealing with large, highly connected datasets, and they need a large amount of labeled data to perform well. The lack of inductive bias in DNNs for relational data further hampers their ability to generalize across tasks where relational information is critical.
      
    In contrast, GCNs are designed to handle graph-structured data, allowing them to represent and process such information naturally. Graph convolutional networks effectively learn and represent node connections, requiring fewer labeled samples and improving their generalization capabilities because of their built-in structure and natural inductive bias for graphs~\cite{kipf2016semi}. Therefore, GCNs offer distinct advantages over DNNs in handling graph-structured data and learning complex relationships.
    However, facial expression recognition using GCNs facial challenges, such as constructing complex graphs, dealing with high-dimensional and irregular data, learning robust features, managing lighting, occlusions, head poses, scalability, and real-time performance. Integrating GCNs with ViTs in facial expression recognition can address challenges by combining GCNs' ability to capture structural relationships between facial landmarks with ViT's strength in modeling both local and global features. This hybrid approach requires advancement in GCNs' architecture and innovative training strategies to effectively merge the strengths of both models, leading to improved accuracy in recognizing complex facial expressions.
    \begin{figure}[!t]
    \centering
    \includegraphics[width=1\linewidth, keepaspectratio]{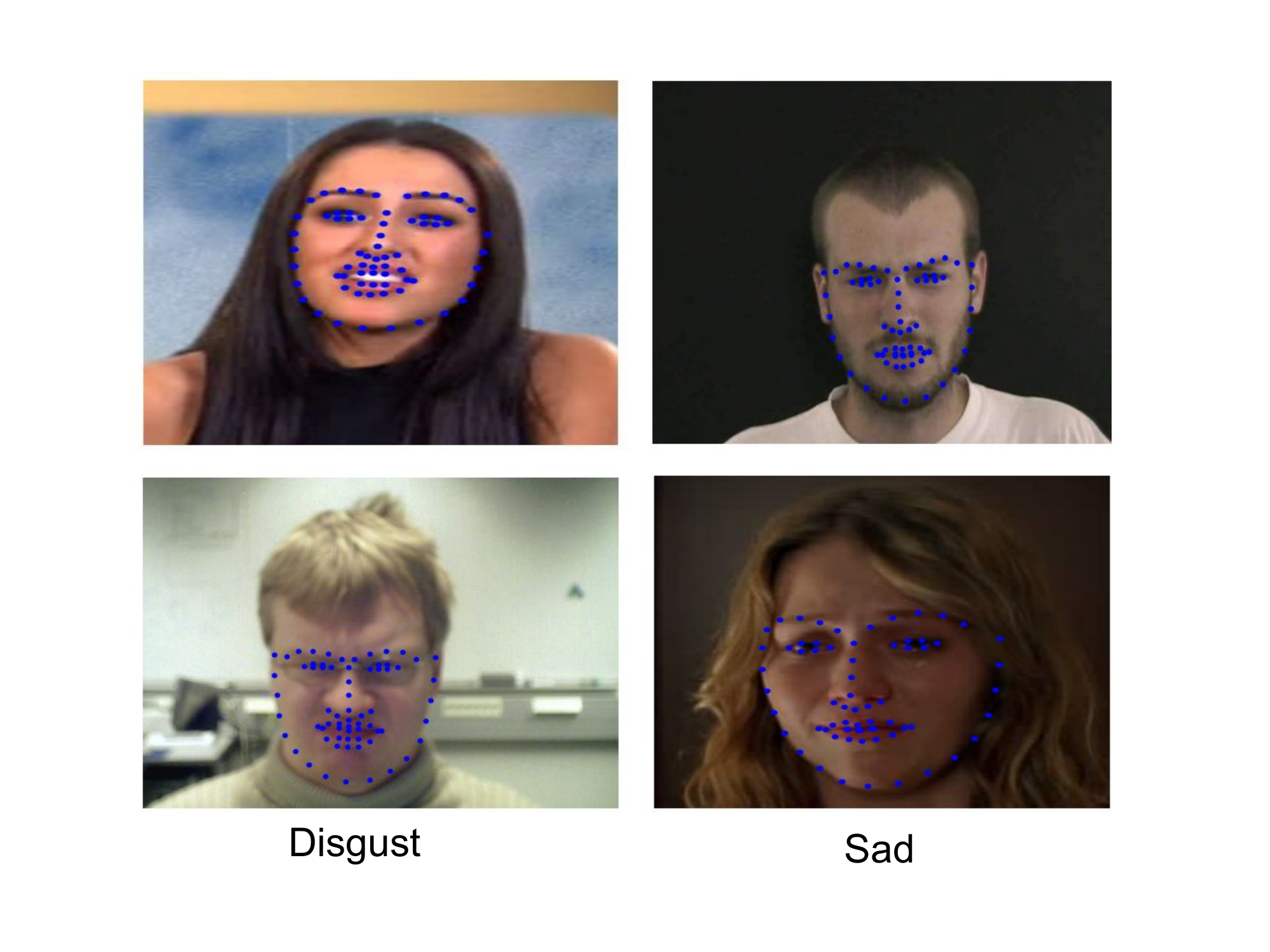}
    \caption{Geometry alone is insufficient. [Best shown in color]}
    \label{fig:geometry}
    \end{figure}     
    Few works explore geometric knowledge of facial attributes for facial expression recognition~\cite{tian2001recognizing,happy2014automatic,kotsia2006facial}. Geometric information, such as relative location and self-deformation, can accurately describe emotional states based on facial observations~\cite{zhao2021geometry}. Geometric face descriptions are more resistant to appearance changes, making them ideal for real-world facial expression recognition applications~\cite{zhang2014facial, zhao2021geometry, zhao2022spatial}. In the GCNs methods~\cite{zhao2021geometry, zhang2014facial, zhao2021geometry, zhao2022spatial, jin2024transformer, xu2024joint, liu2024descriptive, dong2024attentional, al2024ter, mao2025facial, qu2025design, huang2025modeling, tanchotsrinon2011facial, kassab2024gcf, xu2020facial, chen2024dual, ngoc2020facial}, landmarks, AUs (or nodes) and the connections (edges) between them are often predefined or fixed, meaning that the graph's structure remains constant during the learning process. In contrast, our model introduces a more flexible or dynamic approach. Instead of using a fixed graph structure, we are allowing the model to learn the connections between nodes while using the threshold ($\tau$) hyperparameter, potentially evaluating the graph structure during training. Our approach could enable the model to learn more meaningful or relevant relationships between nodes rather than relying on static or predefined connections.

    Our research presents an Exp-Graph framework as shown in Fig.~\ref{fig:pipeline} that uses GCNs~\cite{kipf2016semi,schlichtkrull2018modeling,yao2019graph} to learn geometric descriptions from facial landmarks. The system architecture seeks to increase emotional reasoning from facial images, as geometric information alone cannot distinguish geometrically identical expressions such as disgust and sadness, as shown in Fig~\ref{fig:geometry}. Consequently, GCNs offer a valuable substitute for incorporating the geometric information derived from facial landmarks into emotional representations. Local appearance representations are extracted from landmark positions and aggregated with geometric representations during graph learning. The following briefly describes the primary contributions of this paper:
    \begin{itemize}
      \item Achieves expressive representation of facial expressions by utilizing graph convolutional networks to model structural information extracted from features obtained via a pre-trained vision transformer. 
      \item Captures local and global semantic relationships among facial attributes to learn meaningful expression representations effectively.
      \item Conducts comprehensive evaluations of the proposed Exp-Graph approach on publicly available datasets, including Oulu-CASIA~\cite{zhao2011facial}, eNTERFACE05~\cite{martin2006enterface}, and
     AFEW~\cite{dhall2018emotiw} are publicly available datasets.
    \end{itemize}

   The rest of the paper is organized as follows: Section~\ref{sec:rw} reviews related work in facial expression recognition. Section~\ref{sec:pm} provides a comprehensive overview of the proposed Exp-Graph framework, detailing its design and methodology. Section~\ref{sec:exp} describes the experimental setup, including dataset specifications and evaluation metrics, and presents a thorough analysis of the results. Finally, it concludes the paper and outlines potential directions in \S\ref{sec:con}.

\section{RELATED WORK}\label{sec:rw}
    Since structural information is crucial for facial expression recognition, as emphasized in the previous section, some works have incorporated geometric feature extraction~\cite{kotsia2006facial,song2018geometry,zhang2020geometry} for facial expression recognition. Also, GCNs~\cite{kipf2016semi, lo2020mer, liu2019facial} and vision transformers~\cite{dosovitskiy2020image} have already been explored in different works. Here, we present their limitations and the key differences with our proposed work.
    \subsection{Geometry-based FER}
    Due to the high association between geometric knowledge and expression representations, several researchers propose using landmark geometries for face expression manipulation~\cite{qiao2018geometry,song2018geometry,bodur20213d}. Furthermore, several studies were conducted that suggested using landmark placements as a guide to identify noteworthy local characteristics for representation learning~\cite{mo2020emotion,hasani2017facial,wang2020region}. Kotsia \textit{et al.}~\cite{kotsia2006facial}  employed geometric information to identify informative frames from facial expression sequences, whereas Zhang \textit{et al.}~\cite{zhang1998comparison} incorporated fiducial points on face images to characterize emotions. Gaining momentum from the explosive growth of deep learning technology over the last decade, FER is paying more and more attention to the learning of geometry-associated representations of features. However, most of these techniques adopt multi-task learning approaches instead of directly learning from the geometric data. Devries \textit{et al.}~\cite{devries2014multi} introduced simultaneously learning facial landmark localization and facial expression recognition to enhance the geometric understanding of emotion-related features. Additionally, a multi-domain multi-task network with landmark detection for FER was presented by Gerard and Masip~\cite{pons2018multi}. Zhang \textit{et al.}~\cite{zhang2020geometry} used generative adversarial networks in conjunction with face landmarks to train the pose-invariant features for FER. Investigates using landmark locations as feature descriptions for FER in geometric facial landmarks. However, in practical applications, it is challenging to generate discriminative features due to the poor semantic correlation of these locations. A multimodal auto-encoder was presented by Zhang \textit{et al.}~\cite{zhang2015multimodal} to learn a combined representation from both geometric and visual modalities. Only landmark information is considered in the above research on graph-based representations for extracting geometric information from face images. However, this paper aims to look at a graph-based learning strategy proper for feature representations to provide reliable geometric knowledge.
  
    \subsection{Transformer mechanism for FER}
    Attention mechanisms are explored in the highlight of the regions of interest (RoI) in the domain of facial expression recognition\cite{sun2018visual, li2018patch}. 
    Our studies investigate integrating attention mechanisms into GCNs to enable attentive graph-based representation learning. While transformers, initially proposed for natural language processing (NLP), have become a popular method for handling sequential data, their application in graph-based vision tasks has not been extensively explored. There has been significant research into using ViTs for image interpretation~\cite{ma2021facial,zheng2023poster,zhang2023transformer}. Dosovitskiy \textit{et al.}~\cite{dosovitskiy2020image} introduced a vision transformer for image classification, where an image is divided into patches that serve as tokens to learn non-linear mappings based on dense correlations among all tokens. Yuan \textit{et al.}~\cite{yuan2021tokens} further refined the ViT~\cite{dosovitskiy2020image} approach by developing a more generalized transformer design. Our study aims to extend the vision transformer into the domain of graph-based learning to establish longer-range dependencies among vertices in a series of facial graphs. By doing so, it seeks to enhance the effectiveness of facial expression recognition.
    \subsection{GCNs for Analyzing Facial Images}
    Graph convolutional networks effectively evaluate structured data across various domains, including skeleton-based action detection, NLP, and semi-supervised learning~\cite{kipf2016semi,li2018deeper,zhuang2018dual}. However, while GCNs have been extensively applied to text datasets, their application to image data presents unique challenges that previous studies have not fully addressed.
    In the domain of facial expression recognition, Zhou \textit{et al.}~\cite{zhou2020facial} introduced a FER framework, which uses end-to-end feature learning based on facial topological structures to automatically learn patterns over time and space. However, the method~\cite{zhou2020facial} relies on pre-established facial landmarks identified by HOG as nodes, limiting the flexibility of the graph learning approach. To address the method~\cite{zhou2020facial}, Zhou \textit{et al.}~\cite{zhou2020learning} later introduced an improved method, but it still faced similar limitations.
    Zhao \textit{et al.}~\cite{zhao2021geometry} proposed a geometry-aware FER framework that combines appearance and geometric data using GCNs. This method extensively evaluates the structural information of facial attributes across various expressions and uses CNNs to extract general expression characteristics.
    
    Recently, Qu \textit{et al.}~\cite{qu2025design} combined a spatio-temporal graph convolutional model with a self-attention mechanism, automatically adjusting attention distribution across peak frame images. Luo \textit{et al.}~\cite{luo2024hypergraph} proposed NFER-Former, a hypergraph-guided feature embedding approach designed to model significant facial actions and capture their complex interrelationships, supported by the introduction of a large NIR-VIS facial expression dataset for validation. Dong \textit{et al.}~\cite{dong2024attentional} developed an attention-based visual GCNs for FER, addressing data processing inflexibility by incorporating pixel-level composition strategies. Jin et al. [34] presented a region-of-interest (ROI) -based method, constructing facial graphs from cropped ROIs of action units (AUs) using a deep auto-encoder. Similarly, Chen \textit{et al.}~\cite{jin2024transformer} proposed a node classification approach for FER based on dual subspace manifold learning. Despite these advancements, existing methods cannot dynamically allocate edges and nodes during GCNs training. 
    
     In summary, conventional facial feature extraction methods—such as HOG, LBP, and CNNs—struggle to capture subtle variations and interdependencies in facial expressions effectively. More advanced approaches, including hypergraph-based embeddings and vision transformers, are better equipped to handle these complexities, particularly when applied to large and diverse datasets. Vision transformers, in particular, excel at modeling global context and capturing nuanced feature interactions, making them highly suitable for facial expression recognition. Recent studies have focused on combining vision transformers with graph convolutional networks to improve FER performance. These efforts integrate global visual cues with structural facial information, leverage localized convolutional branches for detailed appearance features, and apply attention mechanisms within graph-based learning frameworks to better understand facial dynamics.

\section{Exp-Graph Framework}\label{sec:pm}
\begin{figure}[!t]
    \centerline{\includegraphics[width=1.0\linewidth, keepaspectratio]{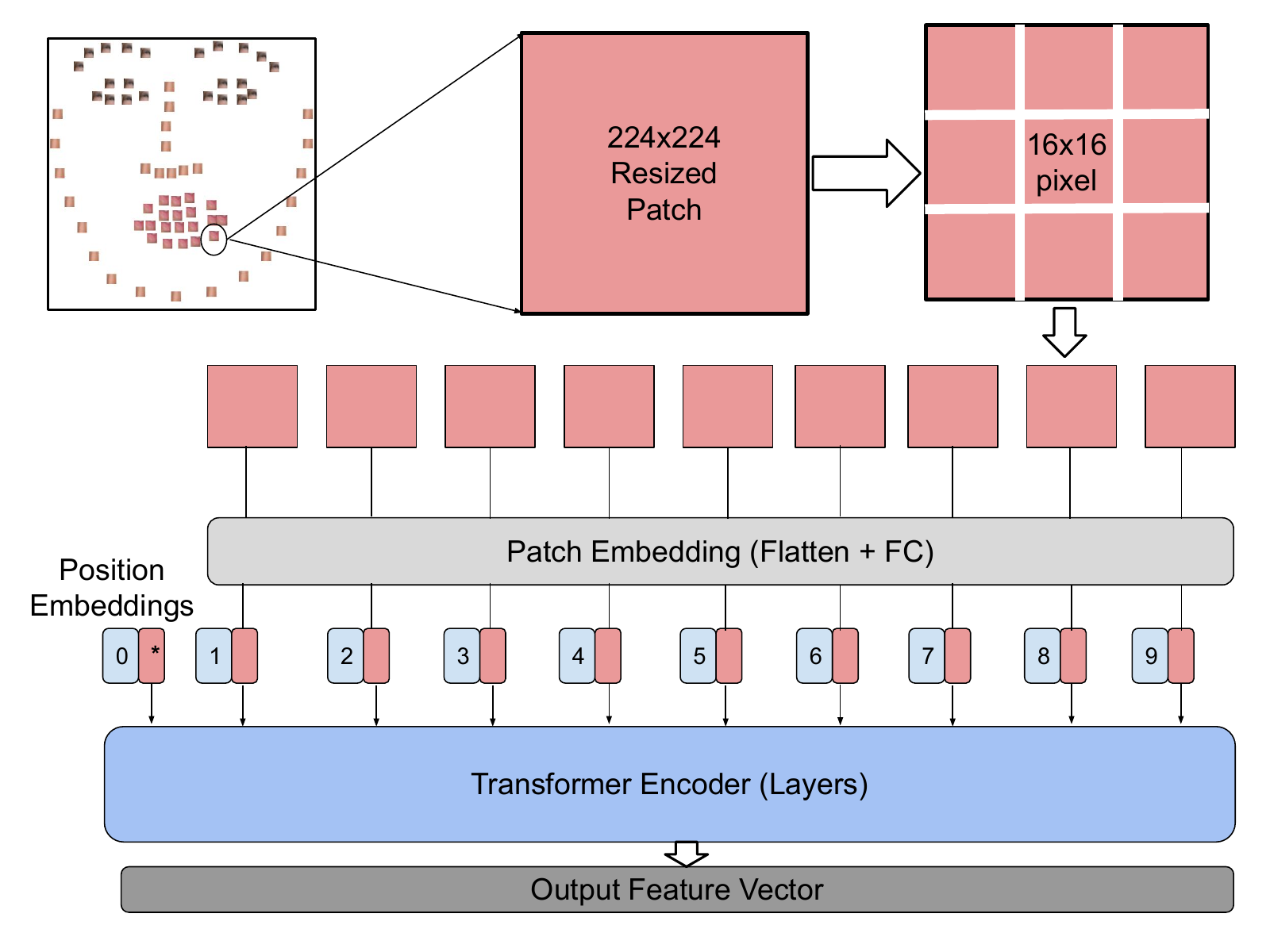}}
    \caption{An illustration of feature extraction of facial units using vision transformer }
    \label{fig:feature_vit}
    \end{figure}

\begin{figure*}[!t]
    \centerline{\includegraphics[width=1.0\linewidth, height=0.7\linewidth]{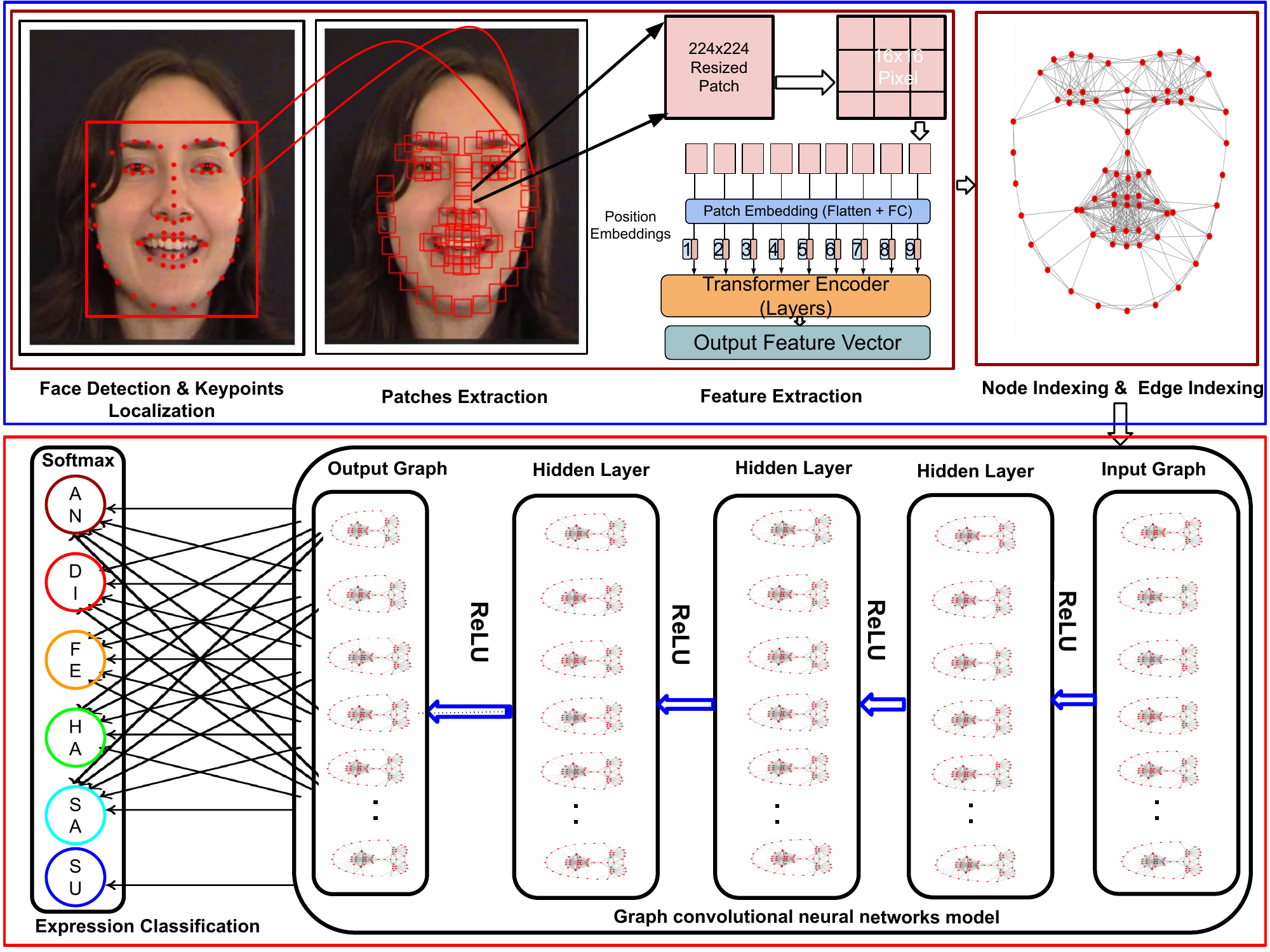}}
    \caption{An outline of the proposed framework for detecting and recognizing facial expressions. The exp-Graph framework is composed of two primary steps: (I) encoding facial attributes through graph-based representation and (II) facial expression recognition through graph convolutional networks. Part (I) is further divided into three subparts: (A) image pre-processing, (B) key points and feature extraction, and (C) graph representation. Part (II) is based on the GCNS model training and facial expression classification. [Best shown in color]}
    \label{fig:architecture}
\end{figure*}

The proposed Exp-Graph framework integrates face detection, feature encoding using pre-trained ViT, as shown in Fig.~\ref{fig:feature_vit}, and recognition via GCNs, as demonstrated in Fig.~\ref{fig:architecture}. The main steps of the framework are as follows: (A) encoding facial attributes through graph-based representation, (B) facial expression recognition through graph convolutional networks. Additionally, we detail the encoding of landmark geometry, the extraction and alignment of local visual features with these geometric representations, and the network architectures involved.
    
    \subsection{Encoding Facial Attributes Through Graph-Based Representation}
   The process begins with image preprocessing to normalize dimensions and improve visual quality. Facial landmarks are detected using the Dlib~\cite{dlib}, and the patch around the landmark points is encoded using a pre-trained ViT~\cite{dosovitskiy2020image}. The facial attribute graph is built by generating an adjacency matrix $\boldsymbol{A}$, which captures relationships between facial landmarks based on their spatial proximity and feature similarity. This approach effectively captures the appearance of facial attributes and relates them to various expressions. The procedure starts by applying $L_2$ normalization to each feature vector, followed by computing a similarity measure $\mathcal{K}(\mathbf{x}_i,\mathbf{x}_j)$ for pairs of feature vectors $\mathbf{x}_i$ and $\mathbf{x}_j$. Simultaneously, a distance matrix is computed based on the squared Euclidean distances between the spatial coordinates of the landmarks. The initial adjacency matrix $\boldsymbol{A_{ij}}$ is derived by normalizing the similarity measure using an exponential function of the Euclidean distance, as illustrated in Eq.~\eqref{eq:equation1}. This method effectively integrates both feature and spatial information into the graph structure.
    \begin{equation}
    \boldsymbol{A}_{ij} = \frac{\mathcal{K}(\mathbf{x}_i, \mathbf{x}_j)}{e^{\| \mathbf{p}_i - \mathbf{p}_j \|}}.
    \label{eq:equation1}
    \end{equation}

    A thresholding step ($\boldsymbol{T}_{s} = \mu_K + \tau \cdot \sigma_K$) is applied to refine the adjacency matrix further. Specifically, a threshold parameter $\tau$ filters out weak connections, retaining only significant relationships between landmarks and $\mu_K$ and $\sigma_K$ are the mean and standard deviation of the $\boldsymbol{A}_{ij}$. As defined in Eq.~\eqref{eq:equation2}, if $\boldsymbol{A}_{ij}$ surpasses the threshold $\boldsymbol{T}_{s}$, it is assigned a value of 1; otherwise, it is set to 0.
     
    \begin{equation}
    \boldsymbol{A} = 
    \begin{cases} 
    1, & \text{if } \boldsymbol{A}_{ij} > \boldsymbol{T}_{s} \\
    0, & \text{otherwise}.
    \end{cases}
    \label{eq:equation2}
    \end{equation}
    
    This formulation ensures that landmarks are close in space and similar features have stronger connections in the matrix. The facial landmarks are nodes, and both feature similarity and spatial proximity weight the edges (connections) between them. The final step applies a threshold to filter out weak connections, leaving only the most significant relationships between landmarks. This refined adjacency matrix can then be used for facial expression recognition for graph construction as shown in the Algorithm~\ref{alg:graph_gen}.
\begin{algorithm}[H]
\caption{Graph Representation}
\label{alg:graph_gen}
\begin{algorithmic}
\STATE \textbf{Input:} Image $\mathbf{I}$
\STATE \textbf{Output:} Graph $\boldsymbol{\mathcal{G}} = (\boldsymbol{P}, \boldsymbol{X}, \boldsymbol{A})$

\STATE $\text{Face} \gets \text{Dlib}(\mathbf{I})$ \hspace{0.5cm} \COMMENT{Detect face in the image}
\STATE $\boldsymbol{P} \gets \text{ExtractLandmarks}(\text{Face})$ \hspace{0.5cm} \COMMENT{Landmark coordinates}

\FOR{each landmark $\mathbf{p}_i$ in $\boldsymbol{P}$}
    \STATE $\mathbf{I}_i \gets \text{ExtractPatch}(\mathbf{I}, \mathbf{p}_i)$ \hspace{0.5cm} \COMMENT{Extract patch around landmark}
    \STATE $\mathbf{x}_i \gets \text{ViT}(\mathbf{I}_i)$ \hspace{0.5cm} \COMMENT{Compute feature vector}
    \STATE Append $\mathbf{x}_i$ to $\boldsymbol{X}$
\ENDFOR

\STATE Compute the adjacency matrix $\boldsymbol{A}$ using Eq.~\eqref{eq:equation1} $\&$ \eqref{eq:equation2}.

\STATE \textbf{return} $\boldsymbol{\mathcal{G}} = (\boldsymbol{P}, \boldsymbol{X}, \boldsymbol{A})$
\end{algorithmic}
\end{algorithm}

\subsection{Facial Expression Recognition Through Graph Convolutional Networks}
The matrix $\boldsymbol{A}$ defines the spatial relationships between facial landmarks, contributing to constructing a graph. We further learn feature embedding for the facial expression recognition using GCNs. The Algorithm~\ref{alg:graph_gen} generates the set of facial landmarks $\boldsymbol{P}$, the set of graph-based features $\boldsymbol{X}$.  

Each layer performs two primary operations:
\begin{enumerate}
    \item \textbf{Node Feature Intergration}: The node features $\boldsymbol{H}^{(l)}$ are integrated based on the graph structure encoded in $\hat{\boldsymbol{A}}$ as in Eq.~\eqref{eq:equation3}.

    \begin{equation}
        \hat{\boldsymbol{A}} \gets \tilde{\boldsymbol{D}}^{-1/2} (\boldsymbol{A} + \boldsymbol{I}) \tilde{\boldsymbol{D}}^{-1/2}.
        \label{eq:equation3}
    \end{equation}
    where $\tilde{\boldsymbol{D}}$ is the degree matrix of $\boldsymbol{A} + \boldsymbol{I}$.
    \item \textbf{Node Latent Feature Projection}: The integrated features are then projected through a learnable weight matrix $\boldsymbol{W}^{(l)}$ and followed by a non-linear activation function $\sigma(\cdot)$ which essential for capturing complex patterns and interactions in the data, allowing the network to learn more expressive and discriminative features in Eq.~\eqref{eq:equation4}.
    \begin{equation}
        \boldsymbol{H}^{(l+1)} \gets \sigma \left( \hat{\boldsymbol{A}} \boldsymbol{H}^{(l)} \boldsymbol{W}^{(l)} \right).
        \label{eq:equation4}
    \end{equation}
    where $\boldsymbol{W}^{(l)}$ is the weight matrix for layer $l$.
\end{enumerate}
\begin{algorithm}[H]
\caption{Exp-Graph Training Model}
\label{alg:gcn_model_algo}
\begin{algorithmic}
\STATE \textbf{Input:} $\boldsymbol{X}\leftarrow$ feature matrix,  $\boldsymbol{A}\leftarrow$ adjacency matrix, $L\leftarrow$ number of layers
\STATE \textbf{Output:} $\boldsymbol{H}^{(L)}\leftarrow$ node feature matrix

\STATE \textbf{Initialization:} Set the initial feature matrix $\boldsymbol{H}^{(0)} \gets \boldsymbol{X}$

\STATE \textbf{Normalize Adjacency Matrix:} Compute the normalized adjacency matrix
\STATE \hspace{0.5cm} $\hat{\boldsymbol{A}} \gets \tilde{\boldsymbol{D}}^{-1/2} (\boldsymbol{A} + \boldsymbol{I}) \tilde{\boldsymbol{D}}^{-1/2}$
\STATE \hspace{0.5cm} where $\tilde{\boldsymbol{D}}$ is the degree matrix of $\boldsymbol{A} + \boldsymbol{I}$.

\FOR{$l = 0$ to $L-1$}
    \STATE \hspace{0.5cm}\textbf{Feature Propagation:} Update the feature matrix
    \STATE \hspace{0.5cm} $\boldsymbol{H}^{(l+1)} \gets \sigma \left( \hat{\boldsymbol{A}} \boldsymbol{H}^{(l)} \boldsymbol{W}^{(l)} \right)$
    \STATE \hspace{0.5cm} where $\boldsymbol{W}^{(l)}$ is the weight matrix for layer $l$.
\ENDFOR

\STATE \textbf{return} $\boldsymbol{H}^{(L)}$
\end{algorithmic}
\end{algorithm}

The Exp-Graph is trained using the cross-entropy loss:
\begin{equation}
\mathcal{L}(Y, \hat{Y}) = - \frac{1}{N} \sum_{j=1}^{N} \sum_{i=1}^{C} y_{ji} \log(\hat{y}_{ji}) 
\label{eq:equation5}
\end{equation} 
where \(\mathcal{L}(Y, \hat{Y})\) denotes the loss function measuring the dissimilarity between the true labels \(Y\) and the predicted probabilities \(\hat{Y}\). \(N\) is the total number of instances (samples) in the dataset. \(C\) is the number of classes. \(Y = [y_{j,i}] \in \{0,1\}^{N \times C}\) is the true label matrix, where \(y_{j,i} = 1\) if the \(j^{\text{th}}\) instance belongs to class \(i\), and \(y_{j,i} = 0\) otherwise. \(\hat{Y} = [\hat{y}_{j,i}] \in [0,1]^{N \times C}\) is the predicted probability matrix, where \(\hat{y}_{j,i}\) is the predicted probability that instance \(j\) belongs to class \(i\). These probabilities are typically obtained via a softmax output layer in multi-class classification. 

\section{Experimental Evaluation}\label{sec:exp}
    This section presents the experimental setup used to evaluate the effectiveness of the proposed approach and compares its performance against existing state-of-the-art (SOTA) methods. We compared our Exp-Graph method with recent and SOTA methods such as DTAGN(Joint)~\cite{jung2015joint}, PPDN~\cite{zhao2016peak}, GCNet~\cite{kim2017deep}, FN2EN~\cite{ding2017facenet2expnet}, DeRL~\cite{yang2018facial}, SASE-FE~\cite{kulkarni2018automatic}, ArcFace + lmrk~\cite{liu2023facial}, ATT~\cite{chen2024dual}, STGCN+AM+PF~\cite{qu2025design}, AT-ViG~\cite{dong2024attentional} on the Oulu-CASIA~\cite{zhao2011facial} dataset. For eNTERFACE05~\cite{martin2006enterface} dataset, we compared with methods such as Mansoorizadeh\textit{et al.}~\cite{mansoorizadeh2010multimodal}, Zhalehpour \textit{et al.}~\cite{zhalehpour2016baum},  Vnet~\cite{zhang2017learning}. For the AFEW~\cite{dhall2018emotiw} dataset, we compared with methods such as HoloNet~\cite{yao2016holonet}, Emotiw2018~\cite{dhall2018emotiw}, DSAN-VGG~\cite{fan2020facial}, DGNN~\cite{ngoc2020facial}. To explore the generalizability of our method, we conducted extensive experiments with the standard benchmark datasets used to evaluate FER using the Exp-Graph method. The datasets and the details used in our experiments are as follows Oulu-CASIA, eNTERFACE05 and AFEW.

    Face detection was performed using Dlib~\cite{dlib}, while the deep learning models were implemented using PyTorch 2.1.2 with CUDA 12.8 support. All experiments were conducted on an NVIDIA RTX A6000 GPU equipped with 48 GB of memory. In our implementation, the hyperparameters are configured as follows: input images are resized to 224×224 pixels. The training begins with an initial learning rate of 0.001, which is gradually reduced to a minimum of 1e-4. A weight decay of 5e-4 is applied to prevent overfitting. The Adam optimizer is employed for optimization, and various activation functions—ReLU, GeLU, and ELU—are explored. The model architecture includes a hidden layer with 256 units and a dropout rate of 0.2 to enhance generalization. To ensure the reproducibility of results, a fixed random seed of 1000 is used. 
    \begin{table}[H]
    \caption{Details of Facial Expression Datasets}
    \label{tab:dataset_details}
    \centering
    \begin{tabular}{|l|c|c|c|}
    \hline
    \textbf{Dataset} & \textbf{Oulu} & \textbf{eNTERFACE05} & \textbf{AFEW} \\ \hline
    \textbf{Resolution}& 320 X 240& 224 X 224 & 224 X 224        \\ \hline
    \textbf{\# Expression} & 6                   & 6                    & 7             \\ \hline
    \textbf{\# Subjects}   & 80                  & 44                   & in-the-wild    \\ \hline
    \textbf{Modality}      & visual              & visual-audio         & visual     \\ \hline
    \end{tabular}
    \end{table}
    Our study used datasets summarized in Table~\ref{tab:dataset_details}. The Oulu-CASIA dataset~\cite{zhao2011facial}, with a resolution of 320x240 pixels and a frame rate of 25 frames per second, captures expressions under three lighting conditions: normal, weak, and dark. The eNTERFACE05  dataset~\cite{martin2006enterface} is significantly larger, containing over 1,200 video sequences from 44 subjects, and is widely used for multi-modality (visual-audio) facial expression recognition and video-based technique evaluation, with each sequence lasting about four seconds and consisting of approximately 120 frames. The AFEW dataset~\cite{dhall2018emotiw}, used in the EmotiW challenge, is a popular video-based FER dataset in the wild, sourced from various television shows and films, presenting challenges such as varying head poses, lighting, and occlusions. 
    \begin{table}[H]
    \caption{Comparison of the Accuracy (\%) of Exp-Graph with the SOTA Methods on Oulu-CASIA Dataset}
    \label{tab:methods_comparison_accuracy_oulu}
    \centering
    \begin{tabular}{|l|c|c|}
    \hline
    \textbf{Method} & \textbf{Accuracy (\%)}  & \textbf{Info.}  \\ \hline
    DTAGN(Joint)~\cite{jung2015joint} & 81.46  & GA+GC  \\ 
    PPDN~\cite{zhao2016peak} & 84.59  & GA  \\ 
    GCNet~\cite{kim2017deep} & 86.11  & GA  \\ 
    FN2EN~\cite{ding2017facenet2expnet} & 87.71  & GA \\ 
    DeRL~\cite{yang2018facial} & 88.0  & GA  \\ 
    SASE-FE~\cite{kulkarni2018automatic} & 89.6  & GA+LA  \\ 
    ArcFace + lmrk~\cite{liu2023facial} & 90.28  & GC+LA  \\ 
    ATT~\cite{chen2024dual} & 89.03 & GC  \\   
    STGCN+AM+PF~\cite{qu2025design} & 90.05 & GC  \\ 
    AT-ViG~\cite{dong2024attentional} & 92.03 & GC  \\\hline 
    \textbf{Exp-Graph (GeLU)} & \textbf{98.09} & GC+LA  \\ 
    \textbf{Exp-Graph (ELU)} & \textbf{98.09} & GC+LA  \\
    \textbf{Exp-Graph (ReLU)} & \textbf{98.09} & GC+LA  \\ 
    \hline
    \end{tabular}\\
    \vspace{2pt}
    \end{table}
    \begin{table}[H]
    \caption{Comparison of the Accuracy (\%) of Exp-Graph with the SOTA Methods on eNTERFACE05 Dataset}
    \label{tab:methods_comparison_accuracy_enter}
    \centering
    \begin{tabular}{|l|c|c|}
    \hline
    \textbf{Method} & \textbf{Accuracy (\%)}  & \textbf{Info.}  \\ \hline
    Mansoorizadeh\textit{et al.}~\cite{mansoorizadeh2010multimodal} & 37.00  & GC  \\ 
    Zhalehpour \textit{et al.}~\cite{zhalehpour2016baum} & 42.12  & GA \\ 
    Vnet~\cite{zhang2017learning} & 54.35  & GA \\ \hline
    \textbf{Exp-Graph (ReLU)} & \textbf{79.01} & GC+LA \\ 
    \textbf{Exp-Graph (GeLU)} & \textbf{79.01} & GC+LA  \\ 
    \textbf{Exp-Graph (ELU)} & \textbf{79.01} & GC+LA  \\ 
    \hline
    \end{tabular}
    \end{table}
    \begin{table}[H]
    \caption{Comparison of the Accuracy (\%) of Exp-Graph with the SOTA Methods on AFEW Dataset}
    \label{tab:methods_comparison_accuracy_AFEW}
    \centering
    \begin{tabular}{|l|c|c|}
    \hline
    \textbf{Method} & \textbf{Accuracy (\%)}  & \textbf{Info.}  \\ \hline
    HoloNet~\cite{yao2016holonet} & 38.81  & GA  \\ 
    Emotiw2018~\cite{dhall2018emotiw} & 38.81  & GA  \\ 
    DSAN-VGG~\cite{fan2020facial} & 52.74  & GA  \\
    DGNN~\cite{ngoc2020facial} & 32.64  & GA+LA  \\
    \hline
    \textbf{Exp-Graph (ReLU)} & \textbf{56.39}  & GC+LA  \\ 
    \textbf{Exp-Graph (GeLU)} & \textbf{56.39}  & GC+LA  \\ 
    \textbf{Exp-Graph (ELU)} & \textbf{56.39}  & GC+LA  \\     
    \hline
    \end{tabular}
    \end{table}

    Table~\ref{tab:methods_comparison_accuracy_oulu},  \ref{tab:methods_comparison_accuracy_enter}, \ref{tab:methods_comparison_accuracy_AFEW} compare the accuracy  of various SOTA methods on Oulu-CASIA, eNTERFACE05 and AFEW datasets, respectively. In particular, our proposed Exp-Graph achieves the highest accuracy of 98.09\%, 79.01\%, and 56.39\% on the Oulu-CASIA, eNTERFACE05, and AFEW datasets, respectively. A comparison of the Exp-Graph with the ensuing methods for facial expression recognition for accuracy, UAR (Unweighted Average Recall), WAR (Weighted Average Recall), cross-entropy loss, and F1-score are presented as metrics for the evaluation. Also, we present results using different thresholds (Th = 0.20, 0.25, 0.30, 0.35, 0.40, 0.45, 0.50, 0.70, 0.90) and patch-size ( 10x10, 20x20, 30x30, 50x50, 70x70, 90x90). Also use the ResNet18~\cite{he2016deep} and EfficientNetB0~\cite{velivckovic2017graph} base model combined with the GAT~\cite{tan2019efficientnet} and GCNs for further exploration in our research with Exp-Graph. 

    \begin{table}[H]
    \caption{Test Performance metrics for the datasets}
    \label{tab:performance_metrics}
    \centering
    \begin{tabular}{|l|c|c|c|c|c|}
    \hline
        \textbf{Metrics} & \textbf{Th0.30} & \textbf{Th0.50} & \textbf{Th0.70} & \textbf{Th0.90} & \textbf{Dataset} \\ \hline
    Loss& 1.34 & 1.21 & 1.65 & 1.74 & Oulu \\
    F1 (\%) & 68.11 & 84.00 & 11.58 & 07.34 & Oulu \\
    WAR (\%) &69.88 & 84.00 & 36.91 & 27.60 & Oulu \\
    UAR (\%)& 85.51 & 92.00 & 86.95 & 87.33 & Oulu \\
    \hline
    Loss& 1.34 & 1.21 & 1.65 & 1.74 & eNTER\\
    F1 (\%) & 68.11 & 84.00 & 11.58 & 07.34 & eNTER \\
    WAR (\%) &69.88 & 84.00 & 36.91 & 27.60 & eNTER \\
    UAR (\%)& 85.51 & 92.00 & 86.95 & 87.33 & eNTER \\
    \hline
    Loss& 1.92 & 1.50 & 1.74 & 1.77 & AFEW\\
    F1 (\%) & 59.27 & 26.95 & 9.03 & 11.06 & AFEW \\
    WAR (\%) &62.67 & 53.54 & 25.71 & 21.08 & AFEW \\
    UAR (\%)& 82.50 & 86.00 & 82.73 & 74.74 & AFEW \\
    \hline
    \end{tabular}  
    \end{table}
    
\begin{figure}[!t]
    \centering

    \subfloat[Oulu-CASIA dataset\label{fig:TestResultsTHwiseOulu}]{%
        \includegraphics[width=1.2\linewidth, keepaspectratio]{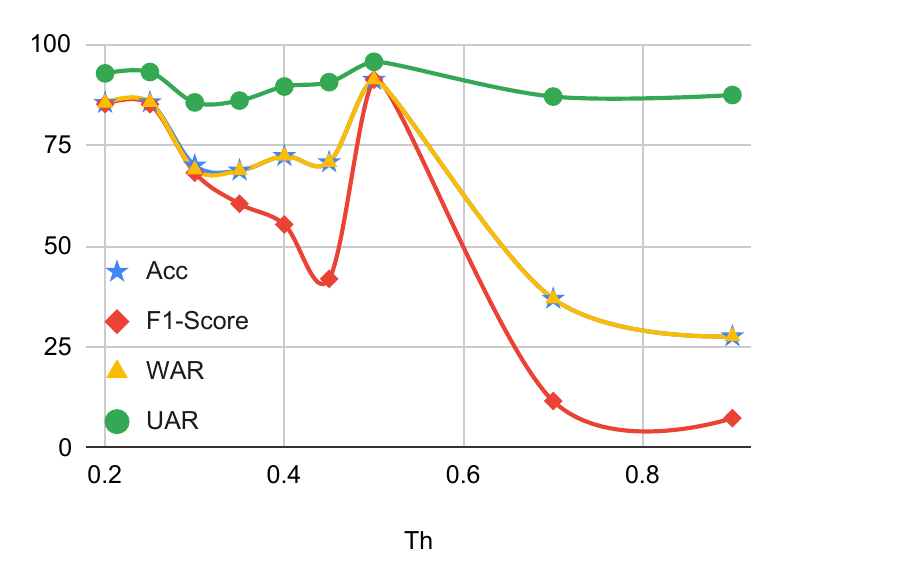}
    }\vspace{1mm}

    \subfloat[eNTERFACE05 dataset\label{fig:TestResultsTHwiseenterface}]{%
        \includegraphics[width=1.2\linewidth, keepaspectratio]{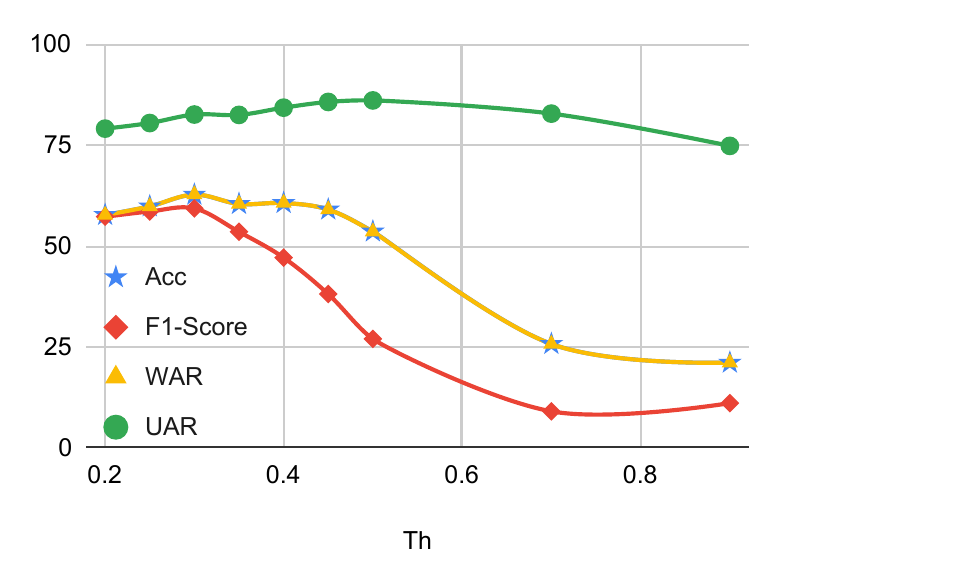}
    }\vspace{1mm}

    \subfloat[AFEW dataset\label{fig:TestResultsTHwiseafew}]{%
        \includegraphics[width=1.2\linewidth, keepaspectratio]{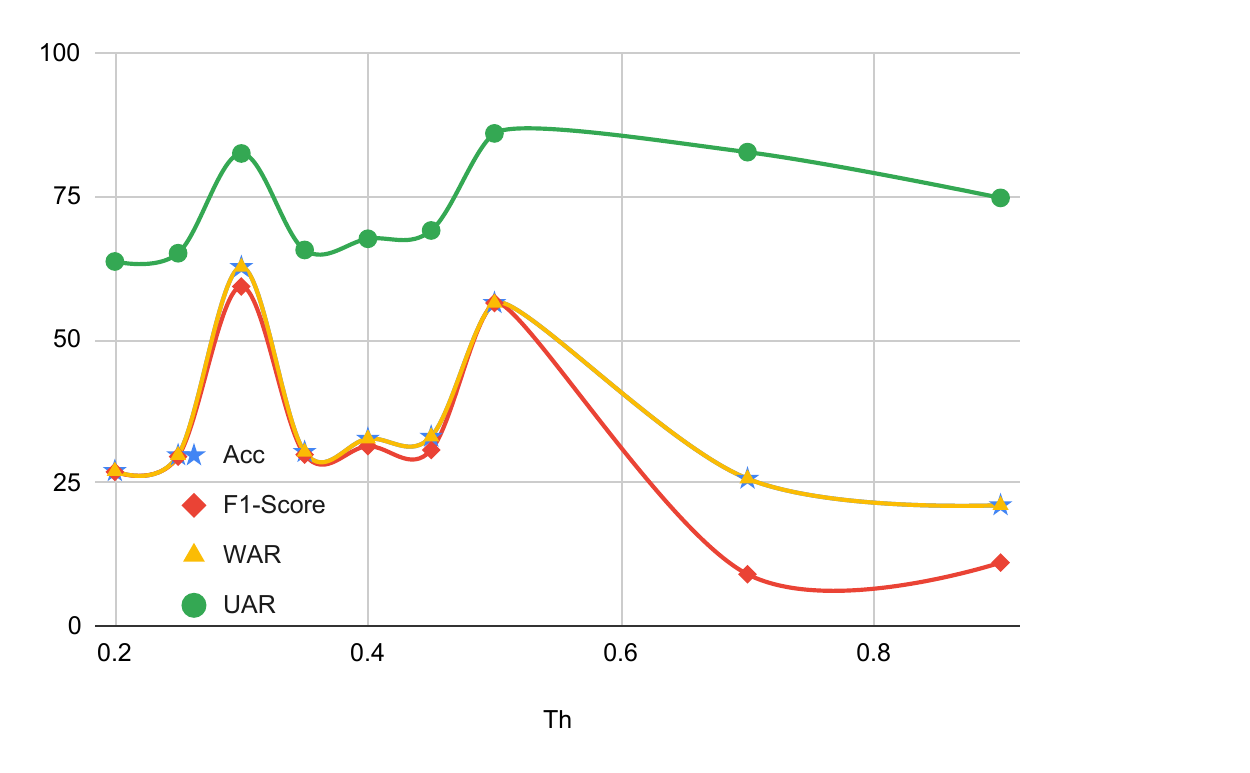}
    }

    \caption{Test results across thresholds (Th = $\tau$) for three datasets. [Best shown in color]}
    \label{fig:TestResultsCombined}
\end{figure}
     
    Table~\ref{tab:performance_metrics} presents the results of analyzing the effect of varying threshold values ($\tau$) on model performance, as illustrated in Figs.~\ref{fig:TestResultsTHwiseOulu}, \ref{fig:TestResultsTHwiseenterface}, and \ref{fig:TestResultsTHwiseafew}. The study evaluates the ViT+GCNs model across three benchmark datasets: Oulu-CASIA, eNTERFACE05, and AFEW. The results show that a threshold of 0.50 consistently yields the best overall performance, particularly in the Oulu-CASIA dataset, where the model achieves an F1 score (84\%), a WAR (84\%), and a UAR (92\%). Performance degrades significantly at higher threshold values, with $\tau = 0.90$ resulting in the poorest results in all metrics on the Oulu-CASIA dataset. A similar trend is observed in the eNTERFACE05 and AFEW datasets, although the threshold of 0.30 yields the worst results in these cases. Notably, the threshold of 0.90 is consistently suboptimal across all datasets, indicating its tendency to introduce excessive sparsity in the relational graphs, which negatively impacts model learning and prediction stability. This configuration benefits from the synergistic use of global appearance (GA) and local appearance (LA) features, outperforming models that rely solely on geometric or individual feature types. The enhanced results can be attributed to the proposed model’s ability to construct and utilize expressive graph representations that capture nuanced structural relationships between facial landmarks.
    
    This inverse relationship between threshold value and performance can be attributed to the increased sparsity and potential noise in the constructed graph at higher thresholds, which results in less informative node connections. Lower thresholds preserve more connections, which are beneficial for learning discriminative patterns, particularly in imbalanced datasets. The accompanying figures—~\ref{fig:TestResultsTHwiseOulu}, \ref{fig:TestResultsTHwiseenterface}, and \ref{fig:TestResultsTHwiseafew}—visually reinforce these findings, showing a clear peak in accuracy and F1 Score around $\tau = 0.50$, followed by a noticeable decline as the threshold increases. While WAR and UAR display slightly more stable trends, they also show reduced performance at higher thresholds. These consistent patterns across all datasets highlight the ViT+GCNs model’s robustness and adaptability, particularly its capacity to handle class imbalance effectively when configured with an appropriately tuned threshold.

\begin{figure}[!t]
    \centering
    \includegraphics[width=1.0\linewidth, height=0.8\linewidth]{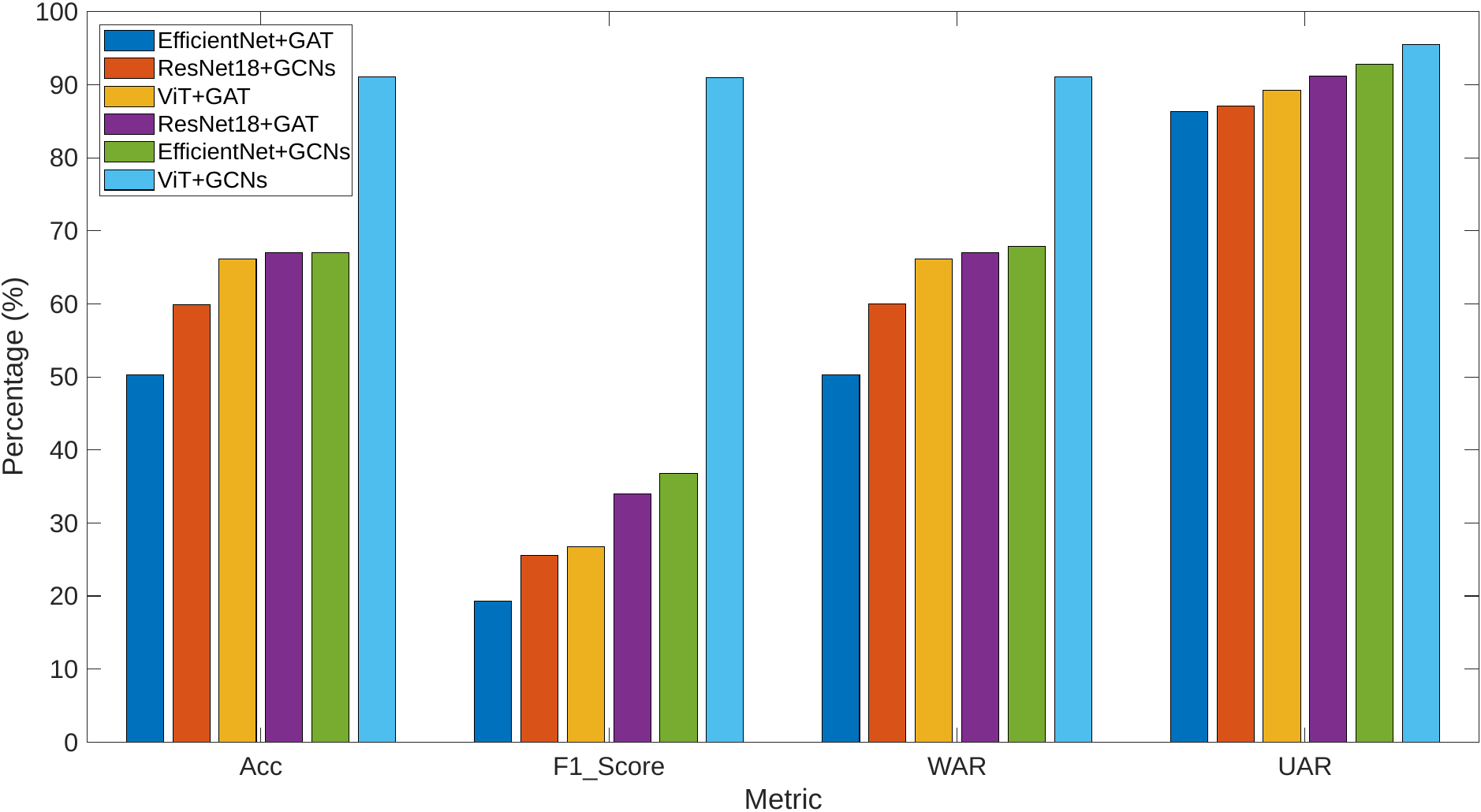}
    \caption{Test result on the Oulu-CASIA dataset across models. [Best shown in color]}
    \label{fig:TestResultsModelwiseOulu}
    \end{figure}

    Fig.~\ref{fig:TestResultsModelwiseOulu} and Table~\ref{tab:Abalationmodel_metrics} present the results comparing the performance of various model architectures on the Oulu-CASIA dataset. Among the evaluated models, the proposed Exp-Graph (ViT+GCNs) configuration consistently achieves the highest performance across the evaluation metrics. These results indicate the model’s superior capacity for accurate classification and its robustness in addressing class imbalance.

    In contrast, models such as EfficientNet+GCNs, ResNet18+GAT, and EfficientNet+GAT demonstrate moderate performance, achieving results that are relatively close to each other but noticeably lower than those of ViT+GCNs. The ViT+GAT configuration yields the weakest performance across all metrics, suggesting that the combination of graph attention mechanisms with vision transformers may not be well-suited to this dataset or task without further architectural refinements. The consistently strong performance of ViT+GCNs highlights the effectiveness of integrating graph convolutional networks with vision transformers, leveraging both spatial relational structure and high-capacity feature extraction. This comprehensive approach to feature extraction and graph-based modeling significantly improves recognition performance, demonstrating the model’s robustness and generalizability on complex facial expressions recognition tasks.
    \begin{table}[!t]  
    \centering
    \caption{Test Results on the Different Models' Performance Metrics on Oulu-CASIA }
    \label{tab:Abalationmodel_metrics}
    \begin{tabular}{|l|c|c|c|c|}
    \hline
        \textbf{Model} & \textbf{ Acc.} &\textbf{F1-Score} & \textbf{ WAR} & \textbf{ UAR} \\
        \hline
        ResNet18+GCNs & 59.93  & 34.00 & 60 & 87.06 \\
        EfficientNet+GCNs & 67.03  & 36.79 & 67.03 & 89.25 \\
        ViT+GCNs & 91.09  & 91.00 & 91.09 & 95.55 \\
        ResNet18+GAT & 67.00  & 26.77 & 67.89 & 92.77 \\
        EfficientNet+GAT &50.30  & 19.29 & 50.30 & 86.28 \\
        ViT+GAT & 66.18  & 25.51 & 66.18 & 91.17 \\
        \hline
    \end{tabular}
    \end{table}

    \begin{figure}[!t]
    \centering
    \subfloat[Oulu-CASIA Dataset (Threshold = 0.20)]{
        \includegraphics[width=1.0\linewidth]{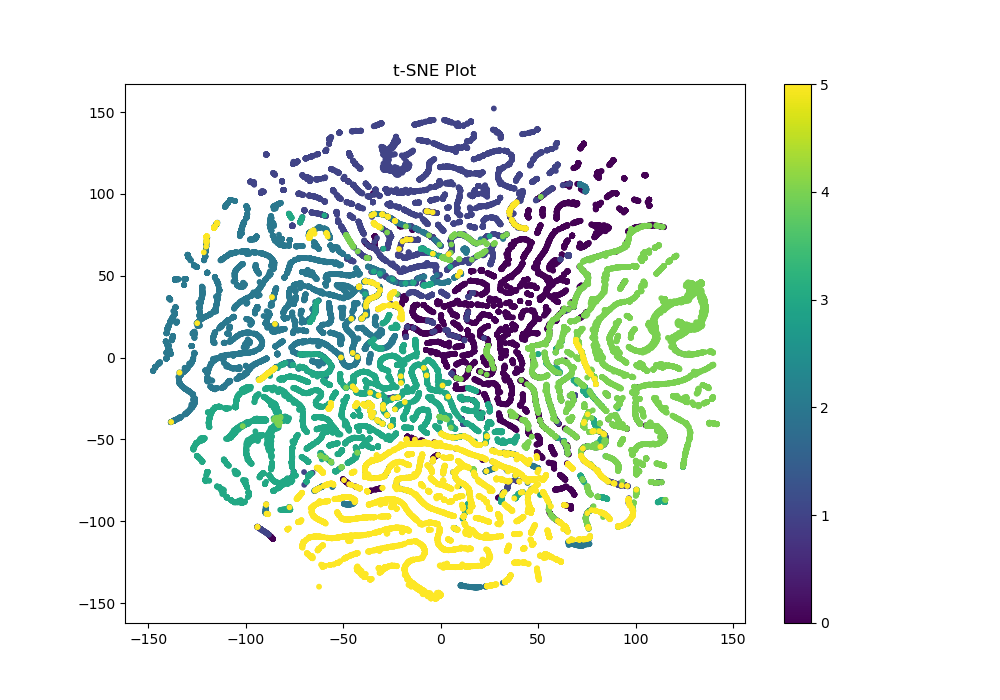}
        \label{fig:t_sne_Oulu}
    }
    \vfill
    \subfloat[eNTERFACE05 Dataset (Threshold = 0.30)]{
        \includegraphics[width=1.0\linewidth]{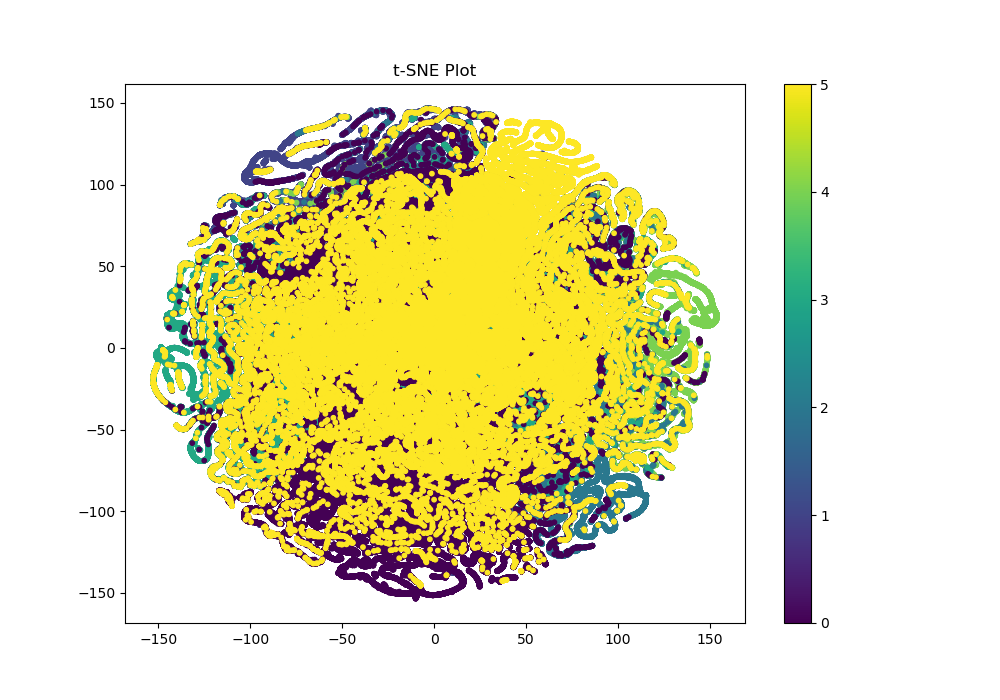}
        \label{fig:t_sne_enterface}
    }
    \caption{t-SNE test results on two FER datasets. [Best shown in color]}
    \label{fig:tsne_combined}
    \end{figure}

    Fig.~\ref{fig:t_sne_Oulu} and Fig.~\ref{fig:t_sne_enterface} present the t-SNE visualizations of learned feature representations on the Oulu-CASIA and eNTERFACE05 datasets, respectively. The plots were generated by evaluating the model’s output at various graph threshold values, specifically $\tau \in {0.20, 0.25, 0.30, 0.35, 0.40, 0.45, 0.50, 0.70, 0.90}$.
    
    Among these, thresholds of 0.20 and 0.30 consistently produced the most distinct and well-separated clusters, indicating effective discrimination between different facial expression classes. Interestingly, the threshold of 0.30 offered good separation, it also resulted in overly dispersed clusters, potentially obscuring the underlying relational structure among expressions. The most balanced and interpretable visualizations were obtained at a threshold of 0.50, where the t-SNE plots showed both clear inter-class separation and coherent intra-class clustering. This suggests that the threshold of 0.20 in the Oulu-CASIA effectively preserves both local and global topological structures within the learned graph representations. The clarity of the clustering at this threshold, particularly when using features extracted from the final layer of the GCNs, underscores its ability to capture complex patterns in facial expression data, thus supporting more accurate and meaningful interpretation of the learned embeddings.
   
    \begin{figure}[!t]
    \centerline{\includegraphics[width=1.0\linewidth, keepaspectratio]{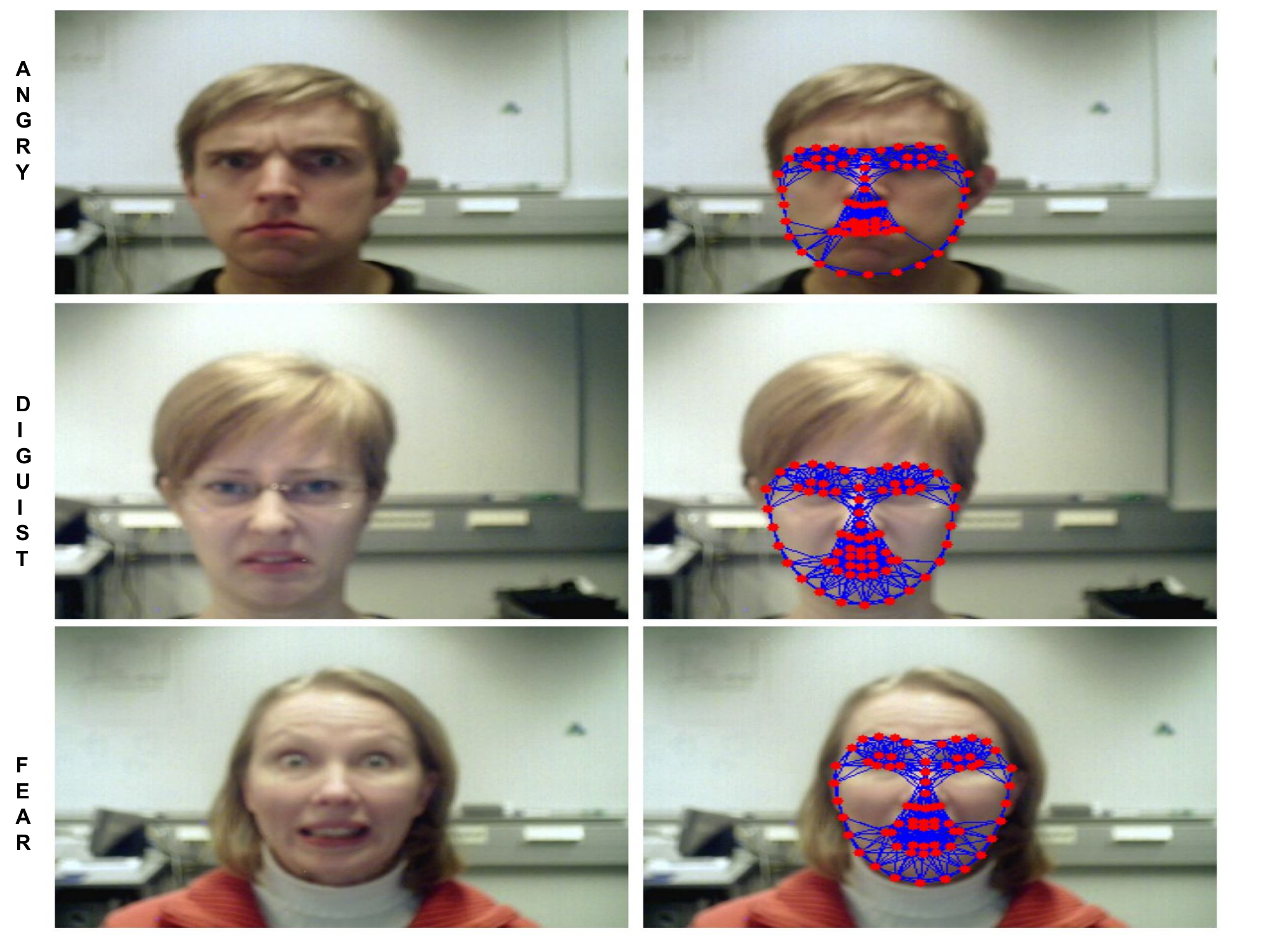}}
    \centerline{\includegraphics[width=1.0\linewidth, keepaspectratio]{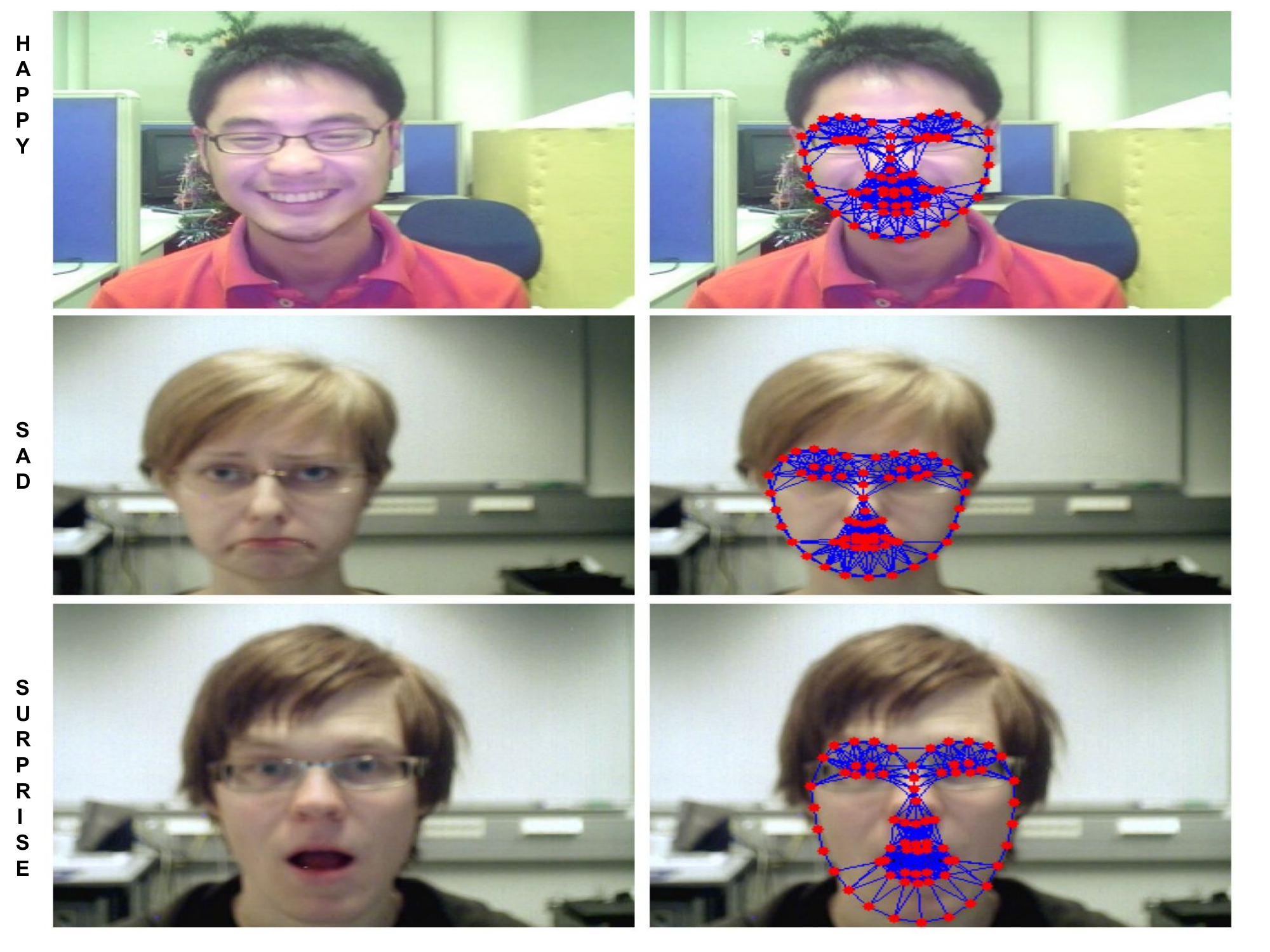}}
    \caption{Visualization of learned graphs for the Oulu-CASIA dataset sample images for $\tau$ = 0.30. [Best shown in color]}
    \label{fig:Connection-Graph}
    \end{figure} 
    
    \begin{table}[!t]
    \centering
    \caption{Performance Metrics for Different Thresholds on the Oulu-CASIA }
    \label{tab:th_metrics}
    \begin{tabular}{|c|c|c|c|c|}
    \hline
    \textbf{Th = $\tau$}  & \textbf{Acc} & \textbf{F1-Score} & \textbf{WAR} & \textbf{UAR} \\ \hline
    0.20    & 85.38 & 85.09 & 85.4  & 92.69 \\
    0.25  & 85.54 & 85.06 & 85.54 & 93.02 \\ 
    0.30   & 69.88 & 68.11 & 68.88 & 85.51 \\ 
    0.35  & 68.64 & 60.4  & 68.64 & 85.96 \\ 
    0.40   & 72.24 & 55.28 & 72.25 & 89.44 \\ 
    0.45  & 70.7  & 41.82 & 70.7  & 90.51 \\ 
    0.50   & 91.09 & 91.1  & 91.1  & 95.55 \\ 
    0.70   & 36.91 & 11.58 & 36.9  & 86.95 \\ 
    0.90   & 27.6  & 7.34  & 27.6  & 87.33 \\ \hline
    \end{tabular}
    \end{table}

    \begin{table}[!t]
    \centering
    \caption{Performance Metrics for Different Thresholds on the eNTERFACE05 }
    \label{tab:th_metrics_v2}
    \begin{tabular}{|c|c|c|c|c|}
    \hline
    \textbf{Th = $\tau$}  & \textbf{Acc} & \textbf{F1-Score} & \textbf{WAR} & \textbf{UAR} \\ \hline
    0.20  & 57.65 & 57.2  & 57.62 & 79.01 \\ 
    0.25  & 59.77 & 58.5  & 59.77 & 80.4  \\ 
    0.30   & 62.67 & 59.27 & 62.7  & 82.5  \\ 
    0.35  & 60.37 & 53.47 & 60.37 & 82.41 \\ 
    0.40   & 60.6  & 47.07 & 60.6  & 84.2  \\ 
    0.45  & 58.98 & 38.05 & 58.98 & 85.62 \\ 
    0.50   & 53.54 & 26.95 & 53.5  & 86    \\ 
    0.70   & 25.71 & 9.03  & 25.7  & 82.73 \\ 
    0.90  & 21.08 & 11.06 & 21.08 & 74.74 \\ \hline
    \end{tabular}
    \end{table}
    
    \begin{table}[!t]
    \centering
    \caption{Performance Metrics for Different Thresholds on the AFEW }
    \label{tab:th_metrics_v3}
    \begin{tabular}{|c|c|c|c|c|}
    \hline
    \textbf{Th = $\tau$}  & \textbf{Acc} & \textbf{F1-Score} & \textbf{WAR} & \textbf{UAR} \\
    \hline
    0.20  & 27.04 & 26.9 & 27.04 & 63.64 \\
    0.25  & 29.81 & 29.56 & 29.81 & 65.08 \\
    0.30  & 62.67 & 59.27 & 62.67 & 82.5 \\
    0.35  & 30.37 & 29.92 & 30.37 & 65.65 \\
    0.40  & 32.69 & 31.38 & 32.69 & 67.6 \\
    0.45  & 33.05 & 30.72 & 33.05 & 69.05 \\
    0.50 & 56.39 & 56.39 & 56.39 & 86 \\
    0.70  & 25.71 & 9.03 & 25.71 & 82.73 \\
    0.90  & 21.08 & 11.06 & 21.08 & 74.74 \\
    \hline
    \end{tabular}
    \end{table}
    Fig.~\ref{fig:Connection-Graph} illustrates the Visualization of the learned graph for the Oulu-CASIA dataset sample images at a threshold value of $\tau = 0.30$. In this visualization, the first column displays samples of different facial expressions (e.g., anger, disgust, fear), while the second column overlays the corresponding connection graph on the original image, highlighting the relational structure among key facial landmarks. The performance metrics of the proposed model across varying threshold values are reported in Tables~\ref{tab:th_metrics}, \ref{tab:th_metrics_v2}, and \ref{tab:th_metrics_v3}, offering insight into the threshold's influence on model effectiveness across different datasets.
    
    Table~\ref{tab:th_metrics}, which presents the performance without data augmentation on the Oulu-CASIA dataset, the model achieves its best performance at $\tau = 0.50$, with an accuracy of 91.09\%, F1-Score of 91.1\%. As the threshold increases (e.g., $\tau = 0.70$ or $\tau = 0.90$), both accuracy and F1-Score degrade noticeably, indicating that excessive sparsity in the graph structure adversely affects learning and classification performance. Similarly, Table~\ref{tab:th_metrics_v2} presents evaluation results on the eNTERFACE05 dataset, demonstrates a similar trend. The highest metrics are recorded around $\tau = 0.30$, with an accuracy of 62.67\% and an F1-Score of 59.27\%, reaffirming the effectiveness of lower thresholds for preserving discriminative information in the relational graph. Finally, Table~\ref{tab:th_metrics_v3} summarizes the model’s performance 
     \begin{figure}
        \centering
    
        \subfloat[Oulu-CASIA dataset (Threshold = 0.3)\label{fig:Patch_oulu}]{%
            \includegraphics[width=1.2\linewidth, keepaspectratio]{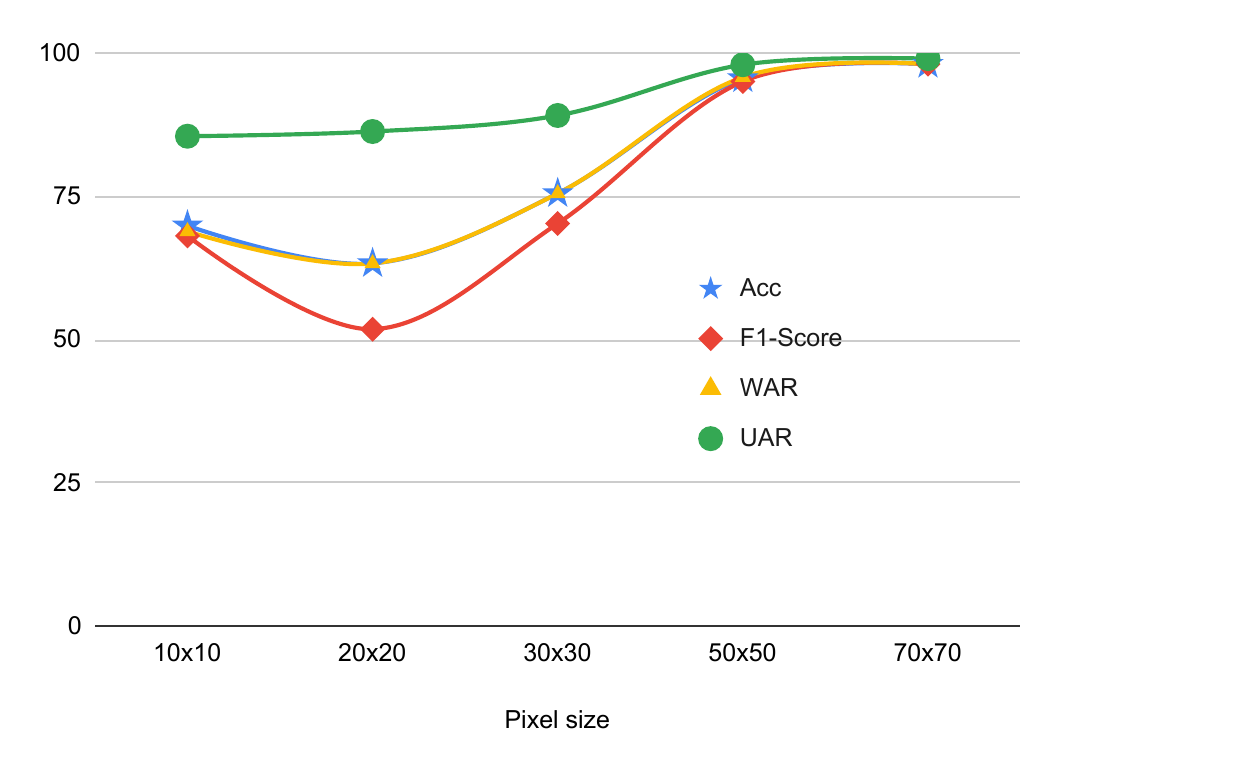}
        }\vspace{1mm}
    
        \subfloat[eNTERFACE05 dataset (Threshold = 0.3)\label{fig:Patch_enterface}]{%
            \includegraphics[width=1.2\linewidth, keepaspectratio]{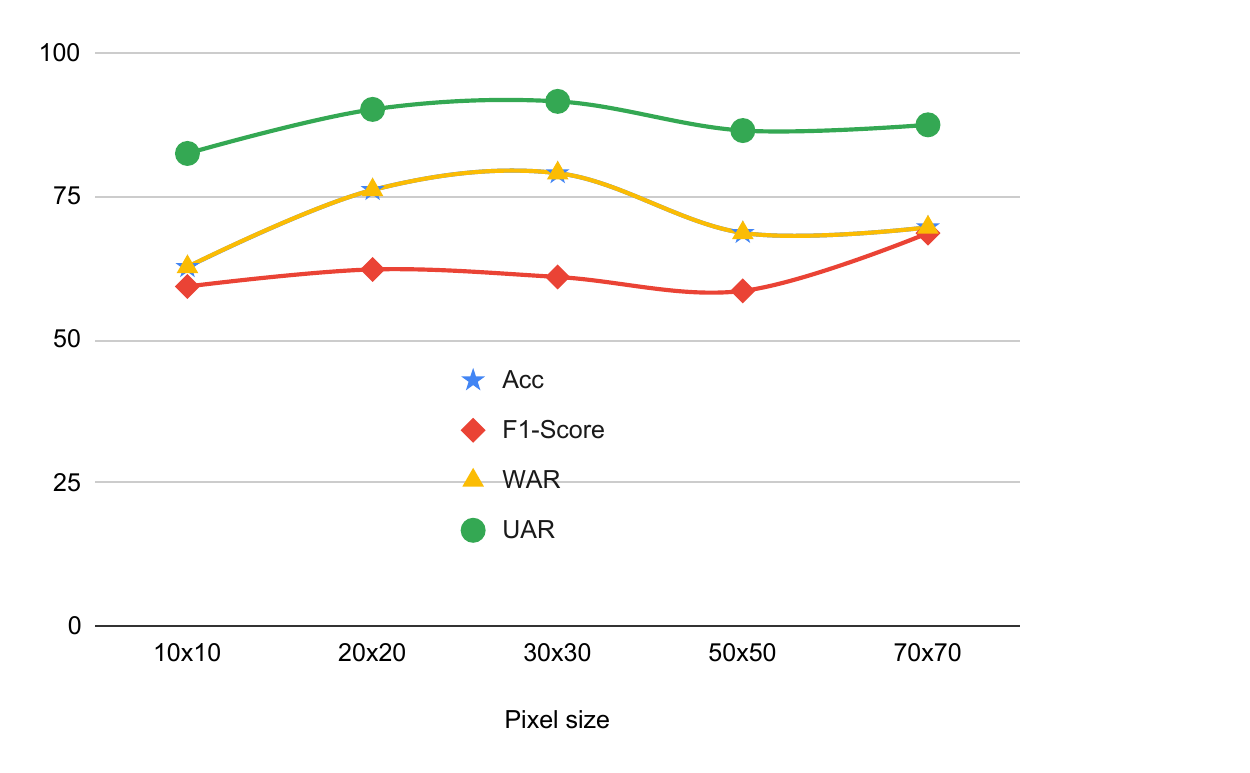}
        }
    
        \caption{Test results across patch sizes for two FER datasets at a threshold of 0.3. [Best shown in color]}
        \label{fig:Patch_combined}
    \end{figure}

    on the AFEW dataset, where the optimal results are also observed at $\tau = 0.30$. These findings collectively highlight the importance of selecting an appropriate threshold value, as it plays a critical role in balancing graph connectivity and expressiveness, ultimately influencing the model’s ability to accurately capture and classify subtle facial expressions across different datasets.
    \begin{table}[!t]
    \centering
    \caption{Performance Metrics for Different Patch sizes on the Oulu }
    \label{tab:Patch_oulu}
    \begin{tabular}{|c|c|c|c|c|}
    \hline
    \textbf{HxW}  & \textbf{Acc} & \textbf{F1-Score} & \textbf{WAR} & \textbf{UAR} \\ \hline
    10x10   & 69.88 & 68.11  & 68.88 & 85.51 \\ 
    20x20  & 63.30 & 51.81  & 63.30 & 86.33  \\ 
    30x30   & 75.52 & 70.27 & 75.52  & 89.13  \\ 
    50x50  & 95.58 & 95.07 & 95.58 & 97.99 \\ 
    \textbf{70x70}  & \textbf{98.09} & \textbf{98.09} & \textbf{98.09} & \textbf{99.06} \\   \hline
    \end{tabular}
    \end{table}    
    
    \begin{table}[!t]
    \centering
    \caption{Performance Metrics for Different Patch sizes on the eNTERFACE05 }
    \label{tab:Patch_enterface}
    \begin{tabular}{|c|c|c|c|c|}
    \hline
    \textbf{HxW}  & \textbf{Acc} & \textbf{F1-Score} & \textbf{WAR} & \textbf{UAR} \\ \hline
    10x10  & 62.67 & 59.27 & 62.7 & 82.5 \\ 
    20x20  & 76.12 & 62.25 & 76.12 & 90.19 \\ 
    \textbf{30x30} & \textbf{79.06} & \textbf{60.93} & \textbf{79.06} & \textbf{91.59} \\  
    50x50  & 68.63 & 58.50 & 68.63 & 86.50 \\  
    70x70  & 69.60 & 68.59 & 69.60 & 87.50 \\    \hline
    \end{tabular}
    \end{table}
    
    \begin{figure}[!t]
        \centering
    
        \subfloat[\footnotesize{t-SNE Test Results on Oulu-CASIA Dataset on the best Patch size.}\label{fig:Patcht_sne_Oulu}]{
            \includegraphics[width=1.0\linewidth]{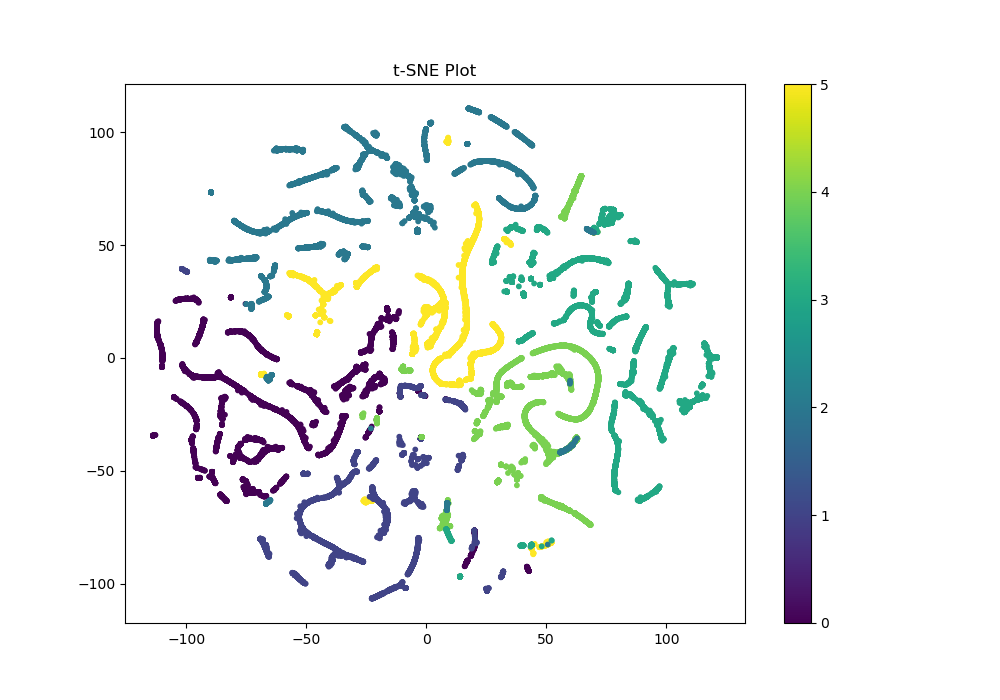}
        }\vspace{1mm}
    
        \subfloat[\footnotesize{t-SNE Test Results on eNTERFACE05 Dataset on the best Patch size.}\label{fig:Patcht_sne_enterface}]{
            \includegraphics[width=1.0\linewidth]{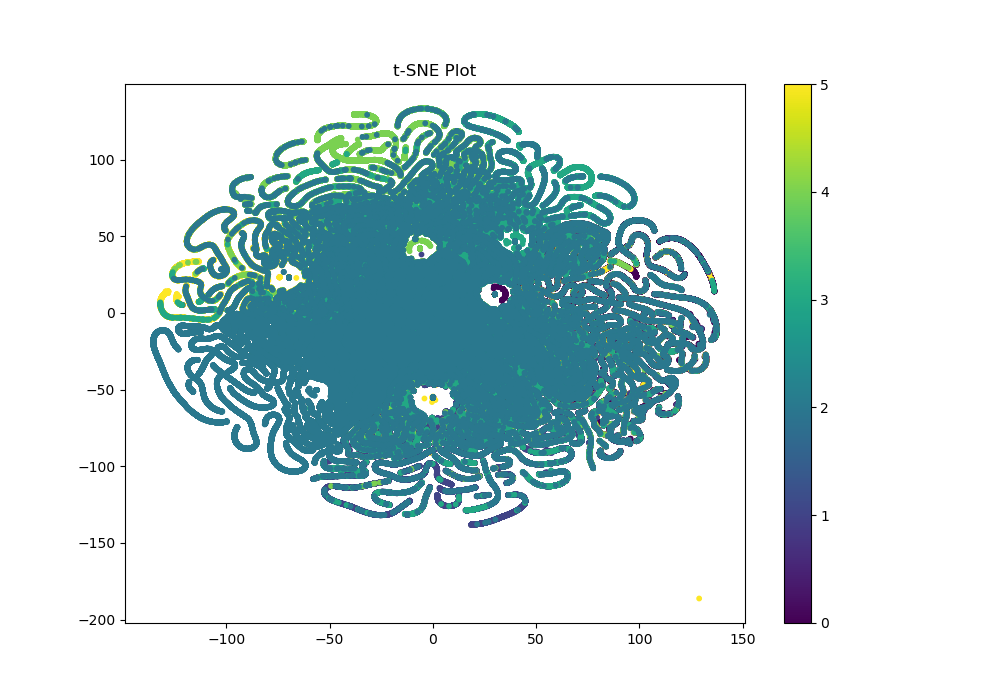}
        }
    
        \caption{t-SNE test results on the best patch size for two datasets. [Best shown in color]}
        \label{fig:tSNE_combined}
    \end{figure} 
  
    As shown in Table~\ref{tab:Patch_oulu} and Fig.~\ref{fig:Patch_oulu}, the performance on the Oulu-CASIA dataset exhibits a positive correlation with increasing patch size. The model achieves its best performance at a patch size of 70×70, obtaining the highest accuracy (98.09\%), F1-score (98.09\%), WAR (98.09\%), and UAR (99.06\%). A noticeable performance improvement begins at the 30×30 patch size, continuing to rise steadily with larger patches, and peaking at 70×70. In contrast, smaller patches such as 10×10 and 20×20 result in significantly lower performance across all metrics, with 20×20 yielding the weakest results, suggesting that smaller patches may fail to capture sufficient contextual and spatial information.

    For the eNTERFACE05 dataset, results reported in Table~\ref{tab:Patch_enterface} and visualized in Fig.~\ref{fig:Patch_enterface}, the optimal performance is observed at a patch size of 30×30, where the model achieves its highest values for accuracy (79.06\%), F1-score (60.93\%), WAR (79.06\%), and UAR (91.59\%). Performance declines for both smaller and larger patch sizes. While 20×20 still performs reasonably well, 10×10 and 50×50 show a marked drop in effectiveness, and the 70×70 patch—which was optimal for the Oulu dataset—fails to deliver comparable results on eNTERFACE05. These differences highlight the dataset-specific sensitivity to patch size and the importance of tailoring patch-based feature extraction to the characteristics of each dataset. Additionally,~\ref{fig:Patch_enterface} presents t-SNE visualizations of the test samples using the optimal patch sizes for each dataset. These plots demonstrate clear and distinct class separations, further validating the effectiveness of the selected patch sizes in preserving discriminative features and supporting accurate classification.
    
    Overall, these results highlight that the optimal patch size varies by dataset, with larger patches proving more effective on Oulu, while a mid-sized patch (30×30) yields the best performance on eNTERFACE05. We also demonstrate how varying the threshold impacts the model’s training and validation performance, recall metrics, and feature detection capabilities. Lower thresholds generally lead to better accuracy, reduced loss, and improved recall, while enabling more comprehensive facial feature analysis.

\section{Conclusion}\label{sec:con}
    In conclusion, this work introduced Exp-Graph, a novel framework for facial expression recognition that utilizes a graph-based representation of facial attribute structures. By modeling facial landmarks as graph nodes and defining edges through spatial and appearance-based relationships, Exp-Graph captures intricate dependencies among facial features. Selecting an appropriate patch size and threshold ($\tau$) is crucial for the optimal performance of the Exp-Graph. An appropriate patch size with a suitable threshold can reduce information loss and result in a more relevant graph representation. In contrast, too large or too small patch sizes and thresholds may result in either similar graph structures or disconnected graphs due to the loss of important information. Therefore, determining the optimal threshold and patch size is essential for effectively preserving facial features and ensuring robust performance. However, the optimal size may vary depending on the characteristics of the dataset. The integration of vision transformers and graph convolutional networks enables the framework to effectively encode global context and structural information. The experimental results confirm the robustness and strong generalization of the method across the datasets.

\bibliographystyle{IEEEtran}
\bibliography{references}
\end{document}